%% file: main.tex
\documentclass[10pt,twocolumn,letterpaper]{article}
\usepackage[pagenumbers]{cvpr} 

\usepackage{graphicx}
\usepackage{wrapfig}
\usepackage{booktabs}
\usepackage{colortbl}
\usepackage{xcolor}
\usepackage{array}
\usepackage{pifont}%
\usepackage{multirow}
\usepackage{marvosym}

\definecolor{MutedGreen}{RGB}{130, 179, 102}
\definecolor{MutedRed}{RGB}{222, 112, 97}
\definecolor{MutedBlue}{RGB}{64, 94, 144}
\newcommand{\cmark}{\ding{51}}
\newcommand{\xmark}{\ding{55}}
\definecolor{myCiteBlue}{HTML}{4682B4}

\usepackage{url}

\newcommand{\thename}{DiffusionDriveV2}

\input{preamble}
\definecolor{cvprblue}{rgb}{0.21,0.49,0.74}
\usepackage[pagebackref,breaklinks,colorlinks,allcolors=cvprblue]{hyperref}
\hypersetup{citecolor=myCiteBlue}
\def\confName{CVPR}
\def\confYear{2026}

\title{DiffusionDriveV2: Reinforcement Learning-Constrained Truncated Diffusion Modeling in End-to-End Autonomous Driving}

\author{
Jialv Zou$^{1,\diamond}$ \quad
Shaoyu Chen$^{3}$ \quad
Bencheng Liao$^{2,1}$ \quad
Zhiyu Zheng$^{4,\diamond}$ \quad
Yuehao Song$^{1,\diamond}$ \\
Lefei Zhang$^{4}$ \quad
Qian Zhang$^{3}$ \quad 
Wenyu Liu$^{1}$ \quad 
Xinggang Wang$^{1,\textrm{\Letter}}$\\
\small\textsuperscript{1} School of Electronic Information and Communications, Huazhong University of Science \& Technology \\ 
\small\textsuperscript{2} Institute of Artificial Intelligence, Huazhong University of Science \& Technology \\ 
\small\textsuperscript{3} Horizon Robotics \quad
\small\textsuperscript{4} School of Computer Science, Wuhan University \\
}

\begin{document}

\maketitle

\let\thefootnote\relax\footnotetext{$^\diamond$ Intern of Horizon Robotics. $^\textrm{\Letter}$ Corresponding author: \url{xgwang@hust.edu.cn}.}

\input{sec/0_abstract}    
\input{sec/1_intro}

\input{sec/2_related}
\input{sec/3_preliminary}

\input{sec/4_method}
\input{sec/5_experiments}

\input{sec/6_conclusion}

{
    \small
    \bibliographystyle{ieeenat_fullname}
    \bibliography{main}
}

\input{sec/X_suppl}

\end{document}

%% file: sec/0_abstract.tex
\begin{abstract}
Diffusion models for trajectory planning in end-to-end autonomous driving often suffer from mode collapse, tending to generate conservative and homogeneous behaviors. While DiffusionDrive employs predefined anchors representing different driving intentions to partition the action space and generate diverse trajectories, its reliance on imitation learning lacks sufficient constraints, resulting in a dilemma between diversity and consistent high quality. In this work, we propose \thename{}, which leverages reinforcement learning to both constrain low-quality modes and explore for superior trajectories. This significantly enhances the overall output quality while preserving the inherent multimodality of its core Gaussian Mixture Model. First, we use scale-adaptive multiplicative noise, ideal for trajectory planning, to promote broad exploration. Second, we employ intra-anchor GRPO to manage advantage estimation among samples generated from a single anchor, and inter-anchor truncated GRPO to incorporate a global perspective across different anchors, preventing improper advantage comparisons between distinct intentions (e.g., turning vs. going straight), which can lead to further mode collapse. \thename{} achieves 91.2 PDMS on the NAVSIM v1 dataset and 85.5 EPDMS on the NAVSIM v2 dataset in closed-loop evaluation with an aligned ResNet-34 backbone, setting a new record. Further experiments validate that our approach resolves the dilemma between diversity and consistent high quality for truncated diffusion models, achieving the best trade-off. Code and model will be available at \url{https://github.com/hustvl/DiffusionDriveV2}
\end{abstract}
\vspace{-0.5cm}

%% file: sec/1_intro.tex
\section{Introduction}
\label{sec:intro}
In recent years, with the growing maturity of traditional tasks such as 3D object detection~\cite{huang2021bevdet,li2024bevformer,yang2023bevformer}, multi-object tracking~\cite{zhang2022bytetrack,zhang2021fairmot}, pre-training~\cite{yang2024unipad,zou2025mim4d,zhang2025visionpad}, online mapping~\cite{liao2022maptr,liao2025maptrv2} and motion prediction~\cite{chai2019multipath,varadarajan2022multipath++}, the development wave in autonomous driving systems has shifted towards end-to-end autonomous driving (E2E-AD), which directly learns a driving policy from raw sensor inputs.

Early approaches in this field have limitations in terms of modeling. Traditional end-to-end unimodal planners~\cite{jiang2023vad,hu2023planning,chitta2022transfuser} regress a single trajectory and fail to propose alternatives for complex driving scenarios with high uncertainty. Selection-based methods~\cite{chen2024vadv2,li2024hydra,li2025hydra} use a large, static vocabulary of candidate trajectories, but this discretization offers limited flexibility.

\begin{figure*}[htbp!]
    \centering
    \includegraphics[width=0.85\linewidth]{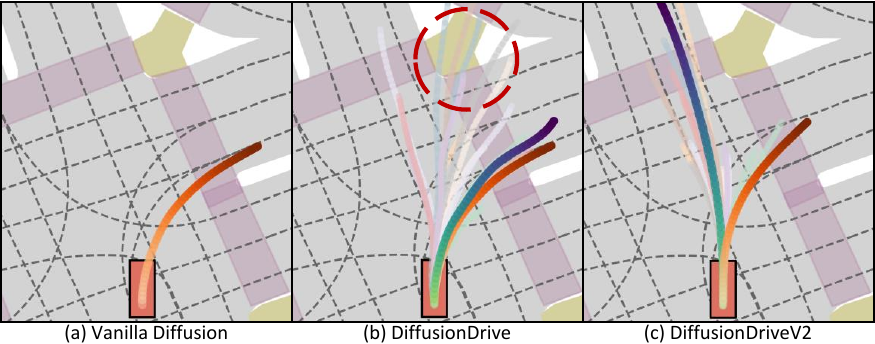}
    \vspace*{-0.3cm}
    \caption{\textbf{Comparison of various models}. (a) \textbf{Vanilla Diffusion} models are prone to mode collapse, collapsing diverse possibilities into a single trajectory. (b) \textbf{DiffusionDrive} generates trajectories with excellent multimodality, yet constrained by imitation learning, it also produces numerous colliding ones (circled in \textcolor{MutedRed}{red}) as most negative modes lack supervision during training, posing a major threat to the system's overall quality. (c) \textbf{\thename{}} leverages reinforcement learning to apply constraints to multi-modal trajectories, guiding the model to generate both diverse and consistent high-quality trajectories.}
    \label{fig:intro}
    \vspace*{-0.5cm}
\end{figure*}

Recently, several approaches have employed diffusion models for trajectory generation~\cite{chi2023diffusion,liao2025diffusiondrive,xing2025goalflow,li2025recogdrive,zheng2025resad}, which can dynamically produce a small set of candidate trajectories conditioned on the surrounding scene. However, directly applying vanilla diffusion models to multi-modal trajectory generation faces the challenge of mode collapse, converging to a single high-probability mode and thus failing to capture the diversity of potential futures, as shown in Fig.~\ref{fig:intro}(a). To address this problem, DiffusionDrive~\cite{liao2025diffusiondrive} proposes constructing the prior distribution of the initial noise using a Gaussian Mixture Model (GMM) defined by multiple predefined trajectory anchors. This structured prior partitions the entire generation space into multiple subspaces, each corresponding to a specific driving intention (e.g., one mode for lane changing, another for driving straight), thereby effectively promoting the generation of diverse behavioral modes.

However, DiffusionDrive faces a fundamental dilemma between the diversity and consistent high quality of its generated trajectories, stemming from its reliance on the imitation learning (IL) paradigm. While the GMM prior enforces diverse mode generation, its training objective, designed to maximize the likelihood of expert trajectories across the entire mixture model, is simplified in practice to optimizing only the parameters of the single positive mode (i.e., the one closest to the expert trajectory). Consequently, it neglects to impose any explicit constraints on trajectories sampled from the negative modes, which constitute the vast majority of the samples. This leads to the model generating high-quality trajectories alongside a multitude of unconstrained, low-quality, and often colliding ones, failing to guarantee consistently high quality, as shown in Fig.~\ref{fig:intro}(b).

This hazardous mix forces reliance on a downstream selector, which is often less robust than the generator due to much fewer parameters. This over-reliance poses a significant risk, as this component is prone to failure when filtering many low-quality trajectories, particularly in out-of-distribution scenarios. 

Reinforcement Learning (RL) provides a powerful solution to this dilemma. In contrast to IL, which is constrained to a single positive mode, RL operates on an \textbf{exploration-constraint} paradigm. On one hand, it raises the model's lower bound by applying goal-alignment constraints to all modes, rewarding desired behaviors while simultaneously penalizing unsafe actions from negative modes. On the other hand, it raises the model's upper bound by pushing the model to explore a broader action space, seeking policies that may exceed the expert's in quality and efficiency. 

Spurred by the success of DeepSeek-R1~\cite{guo2025deepseek}, several works~\cite{li2025recogdrive,song2025breaking} have introduced GRPO to E2E-AD. However, their application has been limited to vanilla diffusion models. Unlike these approaches, in anchored truncated diffusion models, each predefined trajectory anchor represents a distinct driving intention. Naively performing advantage estimation between trajectories corresponding to different driving intentions would exacerbate mode collapse. For instance, trajectories for turning left and going straight should coexist rather than be compared for superiority. This insight motivates us to propose Intra-Anchor GRPO to prevent mode collapse by performing group advantage estimation exclusively within each anchor, thereby blocking comparisons between different intents, and introduce Inter-Anchor Truncated GRPO to provide a global perspective and stabilize training.

With these innovations, we propose a novel framework \textbf{\thename{}}, which leverages RL to address the dilemma between diversity and consistent high quality of DiffusionDrive that stem from its reliance on IL. We benchmark our method on the planning-oriented NAVSIM v1~\cite{dauner2024navsim} and NAVSIM v2~\cite{cao2025pseudo} datasets using closed-loop evaluations. \thename{} sets a new state-of-the-art on both benchmarks, achieving 91.2 PDMS on NAVSIM v1 and 85.5 EPDMS on NAVSIM v2 with ResNet-34 backbone, which represents a substantial improvement over previous methods. Furthermore, compared to other diffusion-based generative models, \thename{} achieves the best trade-off between trajectory diversity and consistent high quality.

Our contributions can be summarized as follows:
\begin{itemize}
    \item We propose \thename{}, a novel approach that introduces RL to address the dilemma between diversity and consistent high quality of DiffusionDrive, which is caused by the incomplete multi-modal supervision in IL. To the best of our knowledge, \thename{} is the first work to directly confront this dilemma and propose a solution.
    \item We introduce Intra-Anchor GRPO and Inter-Anchor Truncated GRPO to solve the issue of inability to perform group advantage estimation across different modes within the Gaussian Mixture Model framework when directly adapting vanilla GRPO to DiffusionDrive. \thename{} is the first work to successfully migrate GRPO to a truncated diffusion model.
    \item We leverage scale-adaptive multiplicative noise as the exploration noise instead of additive noise, which helps preserve the smoothness and coherence of the exploratory trajectories.
    \item Extensive evaluations on the NAVSIM v1 and NAVSIM v2 benchmarks demonstrate that \thename{} significantly improves overall output quality while preserving the ability of the underlying Gaussian Mixture Model to generate multi-modal trajectories, leading to the state-of-the-art performance.
\end{itemize}

%% file: sec/2_related.tex
\section{Related Work}
\label{sec:related}

\begin{figure*}[htbp!]
    \centering
    \includegraphics[width=0.98\linewidth]{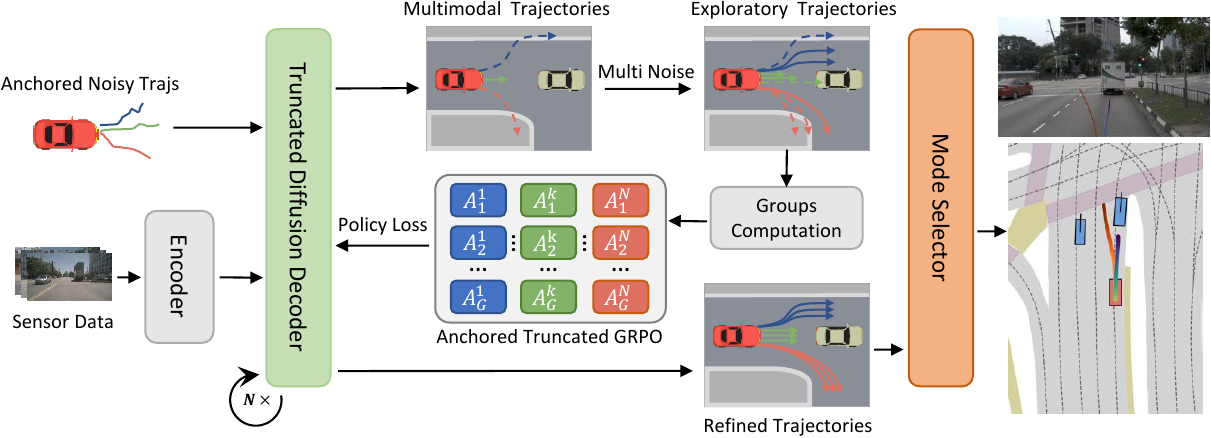}
    \vspace*{-0.2cm}
    \caption{\textbf{Overall architecture of \thename}. Trajectories of different colors represent distinct anchored intents. Solid lines indicate high-quality trajectories, while dashed lines indicate low-quality ones. The truncated diffusion decoder, limited by incomplete supervision in IL, produces low-quality trajectories (\textcolor{MutedBlue}{overtake}, \textcolor{MutedRed}{right turn}) alongside high-quality ones (\textcolor{MutedGreen}{go straight}). To address this, we first apply multiplicative Gaussian noise to push the model to explore the nearby action space. We then propose Anchored Truncated GRPO, which performs intra-group advantage estimation to optimize the model, steering it away from collisions and towards high-quality trajectories. The resulting refined trajectories for \textcolor{MutedBlue}{overtake} and \textcolor{MutedRed}{right turn} become collision-free, while the \textcolor{MutedGreen}{go straight} trajectories become more optimal rather than overly conservative. Finally, a mode selector chooses the most goal-aligned trajectory from the refined trajectories.} 
    \label{fig:arch}
    \vspace*{-0.5cm}
\end{figure*}

\paragraph{\textbf{End-to-End Autonomous Driving.}}
Traditional autonomous driving (AD) systems rely on a highly modular pipeline, which suffers from limitations such as error propagation and information loss between components. UniAD~\cite{hu2023planning} represents a pioneering work that addresses these issues by integrating multiple perception tasks into a single, fully differentiable framework, showcasing the potential of the end-to-end approach. VAD~\cite{jiang2023vad} further improves the system's efficiency by employing a vectorized scene representation. Subsequently, a series of methods~\cite{chen2024vadv2,li2024hydra,li2025hydra,yao2025drivesuprim}, represented by VADv2~\cite{chen2024vadv2} and Hydra-MDP~\cite{li2024hydra},  have shifted to a multi-modal planning framework by performing rule-based scoring and sampling over a fixed vocabulary of anchor trajectories. More recently, Diffusion Policy~\cite{chi2023diffusion} has emerged as a powerful approach for E2E-AD, effectively modeling intricate, multi-modal distributions in high-dimensional action spaces. Diffusion models learn the data distribution through an iterative denoising process and have achieved remarkable performance in image generation tasks~\cite{ho2020denoising,song2020denoising,dhariwal2021diffusion,rombach2022high}. DiffusionDrive~\cite{liao2025diffusiondrive} highlights the challenge of mode collapse for diffusion models in E2E-AD, introduces an anchor-based truncated denoising strategy to counteract it, and drastically improves efficiency by reducing the required denoising steps to just two. 
Overall, diffusion-based generative models show significant potential for E2E-AD, offering both high quality and efficient generation. However, these methods are all fundamentally constrained by the imitation learning paradigm, which means they typically either face the challenge of mode collapse or run the risk of generating numerous low-quality trajectories.

\paragraph{\textbf{Reinforcement Learning for Autonomous Driving.}}  Reinforcement Learning drives an agent to explore by interacting with an environment and maximizing the cumulative return to learn the optimal policy. RL has proven its effectiveness in various domains, from training Large Language Models (LLMs) such as DeepSeek-R1~\cite{guo2025deepseek} and OpenAI O1~\cite{openai2024o1}, to mastering complex games like AlphaGo~\cite{silver2016mastering} and AlphaGo Zero~\cite{silver2017mastering}. Recently, the application of RL in autonomous driving has been increasingly explored. A series of works~\cite{chen2021learning,hu2024solving,lu2023imitation} have investigated the use of RL in non-photorealistic simulators, such as CARLA~\cite{dosovitskiy2017carla}. RAD~\cite{gao2025rad} trains an E2E-AD agent in a realistic 3DGS~\cite{kerbl20233d} environment to bridge the sim-to-real gap. Inspired by the advancements of Deepseek-R1, the application of GRPO has extended to E2E-AD, with AlphaDrive~\cite{jiang2025alphadrive} pioneering its integration into planning reasoning. While subsequent research~\cite{li2025recogdrive,li2025finetuning} have investigated applying GRPO to diffusion-based generative trajectory models, these efforts have been confined to direct implementation on vanilla diffusion, thus suffering from mode collapse.

%% file: sec/3_preliminary.tex
\section{Preliminary}
\label{sec:preliminary}
\paragraph{\textbf{End-to-End Autonomous Driving.}} The E2E-AD system learns an expert driving policy via imitation learning, mapping raw sensor data to future ego-vehicle trajectory predictions. The trajectory is represented by a sequence of future waypoints, denoted as $\tau=\left\{\left(x_{n}, y_{n}\right)\right\}_{n=1}^{N_{f}},$ where $\left(x_{n}, y_{n}\right)$ is the location of each waypoint at time $n$ and $N_{f}$ represents the planning horizon. 

\paragraph{\textbf{Truncated Diffusion Model.}} 
Diffusion policy models~\cite{chi2025diffusion, janner2022planning} learn a reverse Markovian noise process to generate trajectories by iteratively refining a random Gaussian noise. However, experiments show that vanilla diffusion models often suffer from mode collapse, failing to generate diverse driving behaviors. This makes it difficult for them to handle complex driving scenarios and provide a rich set of alternative trajectories, such as car-following versus overtaking, or going straight versus turning left at an intersection.

To overcome the mode collapse problem of vanilla diffusion models, DiffusionDrive~\cite{liao2025diffusiondrive} proposes modeling the trajectory distribution as a Gaussian Mixture Model Distribution. It achieves this by representing a discrete set of driving intents as a set of $N_{anchor}$ anchor trajectories $\left\{\mathbf{a}^{k}\right\}_{k=1}^{N_{\text {anchor }}}$ clustered using K-Means from the expert driving behaviors. Each anchor corresponds to a specific region of the trajectory space, thereby representing a particular driving intent, such as overtaking, turning left, or keeping straight. The trajectory distribution for anchor $\mathbf{a}^{k}$ can be expressed as:
\begin{equation}
p\left(\tau^{k} \mid \mathbf{a}^{k}, z\right)=\mathcal{N}\left(\tau^{k} \mid \mathbf{a}^{k}+\mu^{k}(z), \Sigma^{k}(z)\right).
\end{equation}
Notably, unlike vanilla diffusion models that directly predict trajectories from random noise, DiffusionDrive is trained to predict the offset between a trajectory and its corresponding anchor $\mathbf{a}^{k}$, with $\mu^{k}(z)$ representing a scene-specific offset from the anchor state $\mathbf{a}^{k}$ conditioned on scene context $z$. 
The entire trajectory distribution can be represented as:
\begin{equation} 
\label{eq5}
p(\mathbf{\tau} \mid z)=\sum_{k=1}^{N_{anchor}} s\left(\mathbf{a}^{k} \mid z\right)  p\left(\tau^{k} \mid \mathbf{a}^{k}, z\right).
\end{equation}
This results in a Gaussian Mixture Model (GMM) distribution, where $s\left(\mathbf{a}^{k} \mid z\right)$ is the mixture weight denoting the probability of choosing the driving intent associated with anchor $\mathbf{a}^{k}$, given the scene context $z$.

DiffusionDrive utilizes a truncated diffusion process, which shortens the standard noise schedule to diffuse each anchor trajectory into a corresponding anchored Gaussian distribution:
\begin{equation}
\tau_t^k = \sqrt{\bar{\alpha}_t}\mathbf{a}^k + \sqrt{1-\bar{\alpha}_t}\boldsymbol{\epsilon}, \quad \boldsymbol{\epsilon} \sim \mathcal{N}(0, \mathbf{I}),
\end{equation}
where $t \in [1,T_\text{trunc}]$ and $T_\text{trunc} \ll T$ is the number of truncated diffusion steps. During training, DiffusionDrive takes noisy trajectories $\left\{\tau^{k}_{t}\right\}_{k=1}^{N_{\text {anchor}}}$ as input and predicts denoised trajectories $\left\{\hat{\tau}^{k}\right\}_{k=1}^{N_{\text {anchor}}}$ and the probability scores $\hat{s}^k$, where $s^k$ is a shorthand for $s\left(\mathbf{a}^{k} \mid z\right)$ in Eq.~\eqref{eq5}. 

However, DiffusionDrive is still constrained by the limitations of IL. Although its anchor-based design mitigates mode collapse and provides diverse trajectory options, the training process is fundamentally limited by the fact that only a single GT trajectory is available per scene. Consequently, the model must still select one anchor as the positive mode for optimization during training. 
The anchor closest to the GT trajectory $\tau_{gt}$ is assigned as the positive sample ($y^k$ = 1) and the others as negative samples ($y^k$ = 0). The training objective is:
\begin{equation}
    \mathcal{L} = \sum_{k=1}^{N_\text{anchor}} [y^k \mathcal{L}_{\text{rec}}(\hat{\tau}^k, \tau_\text{gt}) + \mathcal{L}_{\text{BCE}}(\hat{s}^k, y^k)],
\end{equation}
Due to the constraints of IL, only a single mode receives supervision in each scene. As a result, while the model might generate diverse trajectories, it also produces numerous low-quality ones that could lead to collisions, posing a significant hazard to the system.

%% file: sec/4_method.tex
\section{Method}
\label{sec:method}

\subsection{Truncated Diffusion Generator} The overall architecture of our proposed method, \thename{}, is illustrated in Fig.~\ref{fig:arch}. 

To generate multi-modal trajectories, we directly employ DiffusionDrive as our trajectory generator, leveraging its pre-trained weights for IL on GT trajectories. This provides a cold start, equipping our model with an initial capability for multi-modal trajectory generation. Conditioned on features extracted by the perception network, the truncated diffusion decoder takes the noisy trajectories $\left\{\tau^{k}_{t}\right\}_{k=1}^{N_{\text {anchor}}}$ sampled from the anchored Gaussian distribution as input and iteratively refines them over $N_{infer}$ steps to produce the final clean trajectories.

\subsection{Reinforcement Learning for Diffusion Generator}
Although DiffusionDrive demonstrates strong capabilities in generating multi-modal trajectories, it inherits a critical limitation from imitation learning, namely a lack of supervision on negative modes. This often results in the generation of low-quality trajectories, posing a significant threat to the system. To address this, we introduce trajectory-level reinforcement learning objectives to apply constraints across all modes and push the model to explore superior driving policies.
Inspired by DPPO~\cite{ren2024diffusion}, we treat the denoising process as a Markov Decision Process (MDP). Each conditional denoising step in the diffusion chain initiated from anchor $\mathbf{a}^{k}$ is a Gaussian policy:
\begin{align}
\pi_{\theta}\left(\tau_{t-1}^k \mid \tau_{t}^k,z,\mathbf{a}^{k}\right)
    =& \mathcal{N}\left(\tau_{t-1}^k ; \mu_{\theta}\left(\tau_{t}^k, t,z,\mathbf{a}^{k}\right), \right. \nonumber \\ 
    & \qquad \left. \eta\left( 1-\alpha_t \right) I\right),
\end{align}
$\mu_{\theta}\left(\tau_{t}, t,z,\mathbf{a}^{k}\right)$ is the model-predicted mean and $\alpha_t$ is determined by a predefined noise schedule. 

This equation is a Gaussian likelihood, which can be evaluated analytically and is amenable to the policy gradient updates with REINFORCE~\cite{williams1992simple}:
\begin{equation}
\nabla_{\theta} {\mathcal{J}}\left(\pi_{\theta}^k\right)=\mathbb{E}_{\pi_{\theta}^k}\left[\sum_{t=1}^{T_{trunc}} \nabla_{\theta} \log \pi_{\theta}^k\left(\tau_{t-1}^k \mid\tau_{t}^k \right) A_{t}^k\right],
\end{equation}
where $A_t^k$ denotes the advantage function.

\begin{figure}[htbp!]
    \centering
    \includegraphics[width=0.4\textwidth]{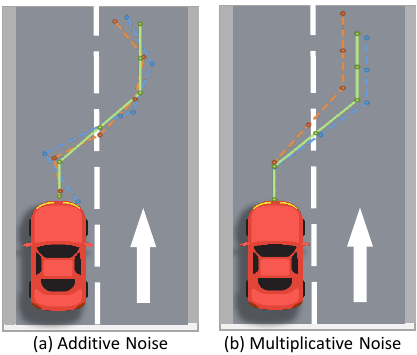}
    \vspace*{-0.2cm}
    \caption{\textbf{Comparison with Different Noise Strategies for Exploration.} The green solid line denotes the original trajectory, while the blue and red dashed lines represent the trajectories after applying exploration noise.}
    \label{fig:multi noise}
    \vspace*{-0.5cm}
\end{figure}

\subsection{Scale-Adaptive Multiplicative Exploration Noise} DiffusionDrive applies the DDIM~\cite{song2020denoising} update rule to drastically reduce the number of denoising steps. This update rule is typically used as a deterministic sampler by setting $\eta = 0$. To enable broader exploration and prevent the issue of calculating a likelihood over a Dirac distribution, we introduce exploration noise by setting $\eta = 1$ during training (equivalent to applying DDPM~\cite{ho2020denoising}), while keeping $\eta = 0$ during validation for deterministic inference. 

However, due to the inherent scale inconsistency between the proximal and distal segments of a trajectory, simply applying additive Gaussian noise at each point disrupts the trajectory's structural integrity and degrades the quality of exploration. As shown in Fig.~\ref{fig:multi noise}(a), this process, where additive Gaussian noise $\epsilon_{add}=\left\{\left(\epsilon_{x,n}, \epsilon_{y,n}\right)\right\}_{n=1}^{N_{f}}$ is applied to a normalized trajectory $\tau=\left\{\left(x_{n}, y_{n}\right)\right\}_{n=1}^{N_{f}}$ typically produces a jagged exploratory path that resembles a broken line, thereby losing its original smoothness. To preserve trajectory coherence, we propose a method that adds only two multiplicative Gaussian noises, one longitudinal and one lateral. It can be expressed as $\tau'=(1+\epsilon_{mul})\tau$, where $\epsilon_{mul} =\left(\epsilon_{long},\epsilon_{lat}\right).$ This scale-adaptive multiplicative noise ensures the resulting exploratory paths remain smooth, as illustrated in Fig.~\ref{fig:multi noise}(b).
\vspace{-0.2cm}
\subsection{Intra-Anchor GRPO for Trajectory Generation} As a reinforcement learning method designed for settings with multiple agents or modes, Group Relative Policy Optimization (GRPO)~\cite{shao2024deepseekmath} updates each agent's policy in relation to a shared, group-level baseline. This approach diverges from conventional PPO~\cite{schulman2017proximal} by defining the policy gradient through an advantage function that is normalized by group-conditioned expectations. By optimizing for non-differentiable objectives with trajectory-level rewards, GRPO enhances standard imitation learning, guiding the diffusion model to produce diverse, goal-oriented trajectories with the potential to surpass human-driver performance.

However, naively using trajectories sampled from different anchors as the ``group'' for the GRPO policy update would be counterproductive. This approach contradicts our core motivation for using anchors to partition the trajectory space into distinct regions that correspond to different driving intentions, and it would even lead to mode collapse. For instance, if samples from anchors representing 'turn right' and 'go straight' (as shown by the red and green trajectories in Fig.~\ref{fig:arch}) were optimized relative to each other, the policy would likely collapse to the single, more common 'go straight' mode. These anchors represent fundamentally different intents and should not be directly compared within the same optimization group.

Based on this insight, we propose Intra-Anchor GRPO. For each anchor, we first generate a group of G trajectory variations by diffusing the anchor with random Gaussian noises and exploration noise. We then perform the GRPO update within this group of G trajectories, rather than across groups from different anchors. This approach constrains the policy optimization to the state space of each specific behavioral intent, guiding the model to generate safer and more goal-oriented trajectories without compromising its multi-modal capabilities. The RL loss function can be represented as:
\begin{align}
L_{RL}=&-\frac{1}{N_{anchor}} \sum_{k=1}^{N_{anchor}}\frac{1}{G} \sum_{i=1}^{G} \frac{1}{T_{trunc}} \sum_{t=1}^{T_{trunc}} \nonumber \\
& \quad \gamma_{t-1} \log \pi_{\theta}\left(\tau_{t-1}^{k,i} \mid \tau_{t}^{k,i}\right) A^{k,i},
\label{eq7}
\end{align}
where $\gamma_{t-1}$ is the discount coefficient mitigating instability in early denoising steps, and $A^{k,i}$ is the advantage function, which GRPO estimates by computing the group-relative advantage, thereby avoiding the need for a value model. 
\begin{equation}
A^{k,i}=\frac{r^{k,i}-\operatorname{mean}\left(\left\{r^{k,1}, r^{k,2}, \cdots, r^{k,G}\right\}\right)}{\operatorname{std}\left(\left\{r^{k,1}, r^{k,2}, \cdots, r^{k,G}\right\}\right)},
\end{equation}
denotes the group relative advantage with $r^{k,i} = R(\tau_0^{k,i})$. A single reward estimate $R(\tau_0^{k,i})$, calculated from the final clean trajectory $\tau_0^{k,i}$, is applied to all denoising steps in the diffusion chain, with the influence of each step being scaled by denoising discount $\gamma_{t-1} \in\left(0,1\right)$.

Furthermore, analogous to how the original GRPO provides regularization by adding a KL divergence between the policy model and the reference model, we incorporate an additional imitation learning loss to prevent the model from overfitting and potentially compromising its general driving capabilities. The combined loss is
\begin{equation}
L = L_{RL} + \lambda L_{IL},
\end{equation}
with weight coefficient $\lambda \in\left(0,1\right)$.

\subsection{Inter-Anchor Truncated GRPO for Trajectory Generation} 
While Intra-Anchor GRPO prevents mode collapse, completely isolating the modes introduces a new problem: the advantage estimates lose global comparability. 
For example, a suboptimal but safe trajectory in one mode could get a negative advantage, while a dangerous, colliding trajectory in another mode might get a positive advantage if it's the 'best' sample within its own group. This reliance on local, intra-group comparisons can provide a misleading learning signal to the model.

To address this issue, we introduce Inter-Anchor Truncated GRPO, which is guided by a simple yet powerful principle: reward relative improvements, but only penalize absolute failures. We implement this by modifying the advantage estimates $A^{k,i}$ from Intra-Anchor GRPO. Specifically, we truncate all negative advantages to zero and assign a strong penalty of -1 to any trajectory that results in a collision:

\begin{equation}
A_{trunc}^{k,i} =
\begin{cases}
-1  & \text{if collision,} \\
\max(0, A^{k,i})  & \text{otherwise.}
\end{cases}
\end{equation}

This provides the model with a clear and consistent learning signal. This truncated advantage $A_{trunc}^{k,i}$ subsequently replaces $A^{k,i}$ in the RL loss calculation (Eq.~\eqref{eq7}).

\subsection{Mode Selector} We append a final mode selector to our model, which is responsible for choosing the optimal, goal-aligned trajectory from multi-modal predictions representing distinct intents. A higher score corresponds to a stronger alignment with the overall objective. 
Specifically, the trajectory coordinates serve as queries, first interacting with BEV features via deformable spatial cross-attention, and then being refined by cross-attention layers with agent and map queries. Finally, the context-rich representation is passed to a Multi-Layer Perceptron (MLP) to predict the score. Inspired by DriveSuprim~\cite{yao2025drivesuprim}, we employ a two-stage, coarse-to-fine scorer. The process begins with a coarse scorer that initially selects the top-k candidate trajectories, which are subsequently passed to a fine-grained scorer for a more detailed selection. The score learning uses the binary cross-entropy (BCE) loss.

For the continuous metric, we introduce an additional Margin-Rank loss: 
\begin{equation}
\mathcal{L}_{\text {rank }}=\frac{1}{N} \sum_{i, j} \max \left(0,-\operatorname{sign}\left(s_{i}-s_{j}\right) \cdot\left(\hat{s}_{i}-\hat{s}_{j}\right)+m\right),
\end{equation}
where $s$ are the ground truth, while $\hat{s}$ are predicted scores. The margin $m$ is a positive hyperparameter. 
This loss guides the model to compare the relative quality of trajectories, which avoids the difficulty of directly regressing their absolute continuous values. As a result, the model's ability to discriminate between subtle differences is further enhanced.

%% file: sec/5_experiments.tex
\section{Experiments}
\subsection{Benchmark}
\paragraph{\textbf{Dataset.}} We evaluate \thename{} on the NAVSIM v1~\cite{dauner2024navsim} and NAVSIM v2~\cite{cao2025pseudo} datasets.
NAVSIM offers a collection of real-world, planning-centric driving scenarios built upon OpenScene~\cite{contributors2023openscene}, which is a compact version of the extensive nuPlan~\cite{caesar2021nuplan} dataset. It features data from a sensor suite combining eight cameras for a 360° field of view (FOV) with a merged point cloud from five LiDARs. The dataset is split into \texttt{navtrain} (1,192 training scenes) and \texttt{navtest} (136 evaluation scenes).

\input{tables/navsim}

\subsection{Implementation Details}
To ensure a fair comparison, our model employs the same ResNet-34~\cite{he2016deep} backbone as Transfuser and DiffusionDrive, while also matching the diffusion decoder size used in DiffusionDrive. \thename{} takes three cropped and downscaled forward-facing camera images, concatenated as a 1024$\times$256 image, and a rasterized BEV representation of the LiDAR point cloud as input. We use the pre-trained weights from DiffusionDrive as a cold start. Our model is then trained for 10 epochs on the \texttt{navtrain} split using reinforcement learning. We use the AdamW optimizer with a learning rate of $2\times 10^{-4}$ and a total batch size of 512, distributed across 8 NVIDIA L20 GPUs. The mode selector was trained for 20 epochs using the same configuration. In inference, similar to DiffusionDrive, our model can generate predictions using only 2 denoising steps. Detailed hyperparameter settings and dataset metric details are provided in the supplementary material.

\subsection{Main Results}
\paragraph{Result on NAVSIM v1.} As shown in Tab.~\ref{tab:navsim}, \thename{} achieves a state-of-the-art performance on NAVSIM v1 \texttt{navtest} split, with a PDMS of 91.2. Our model outperforms DiffusionDrive by 3.1 PDMS and significantly boosts the EP score by 5.3, which demonstrates that our model provides higher-quality and more efficient driving strategies. This improvement is attributed to our carefully designed reinforcement learning method. Compared to DIVER, which is also based on reinforcement learning, \thename{} also surpasses it by 2.9 PDMS. This result demonstrates the superior effectiveness of our Intra-Anchor GRPO and Inter-Anchor Truncated GRPO training framework. Moreover, \thename{} equipped with only a ResNet-34 backbone (21.8M params) still outperforms GoalFlow and Hydra-MDP, which are built upon the larger V2-99~\cite{lee2020centermask} backbone (96.9M params).
\vspace{-0.2cm}
\paragraph{Result on NAVSIM v2.} \thename{} maintains its strong performance on the more challenging NAVSIM v2 dataset, achieving a new state-of-the-art EPDMS as shown in Tab.~\ref{tab:navsimv2}. To ensure a fair comparison, all models evaluated in this benchmark utilize the same ResNet-34 backbone.
\input{tables/navsimv2}

\paragraph{Diversity and Quality.} 
Inspired by DIVER~\cite{song2025breaking}, we introduce a \textbf{Diversity Metric} (denoted as \textbf{$Div.$}) to quantitatively evaluate a model's ability to generate multi-modal trajectories. The metric defines the unnormalized pairwise diversity at waypoint $n$ as:
\begin{equation}
\label{eq:div_sw_raw}
Div_{\text {raw }}^{n}=\frac{2}{M(M-1)} \sum_{i=1}^{M-1} \sum_{j=i+1}^{M}\left(\left\|\mathbf{p}^{i}_n-\mathbf{p}^{j}_n\right\|_{2}\right).
\end{equation}
To ensure scale consistency for trajectories across different scenarios, they are normalized by the average trajectory scale:
\begin{equation}
\label{eq:div_sw_norm}
Div^n=\min \left(1, \frac{Div^{n}_{\mathrm{raw}}}{\epsilon+\frac{1}{M} \sum_{m=1}^{M}\left\|\mathbf{p}^{m}_n\right\|_{2}}\right).
\end{equation}
We report the average $Div^n$ across all waypoints as the final $Div.$ score. To assess the overall quality of the generated trajectories, we further report the Top-K PDMS. Since regression-based and selection-based E2E-AD methods can only generate deterministic trajectories, we exclusively compare our approach against other diffusion-based methods in the \texttt{navtest} dataset. Following DiffusionDrive, each model generates 20 trajectories for evaluation. The results are presented in Tab.~\ref{tab:div}.
\input{tables/div}

The results presented here are the models' raw outputs, evaluated before their respective selection modules (i.e., DiffusionDrive's classifier and \thename{}'s selector). The goal of this analysis is to determine the extent to which these models produce low-quality trajectories and depend on their selection modules to ensure high quality. This over-reliance is a critical concern, as selector modules typically have fewer parameters, leading to weaker generalization capabilities. Consequently, they are prone to failure in out-of-distribution scenarios, creating a significant hazard for the system.

The results validate our theory. Vanilla diffusion methods achieve consistent generation quality but lack diversity, collapsing into a single ``bored'' trajectory. DiffusionDrive achieves very high generation diversity but fails to guarantee consistent high quality. In contrast, \thename{} utilizes a meticulously designed RL algorithm to achieve an exploration-constraint effect. It imposes constraints on all modes, raising the model's lower bound (Top-10 PDMS), and pushes the model to explore better policies, raising its upper bound (Top-1 PDMS). \thename{} resolves the dilemma between diversity and consistent high quality for truncated diffusion models, achieving the best trade-off.

\subsection{Ablation Studies}
We conduct a series of ablation studies to verify the effectiveness of each design in \thename{}. In this section, all ablation studies are conducted with the same set of hyperparameters for a fair comparison and are trained for fewer epochs for rapid validation.
\paragraph{The Type of Exploration Noise.} Tab.~\ref{tab:abl1} shows the results of using different exploration noises. The findings confirm that scale-adaptive multiplicative noise is superior to additive noise, as it effectively addresses the scale inconsistency between the proximal and distal parts of a trajectory.

\input{tables/abl_1}

\paragraph{Impact of Intra-Anchor GRPO.} Tab.~\ref{tab:abl_intra} shows the impact of Intra-Anchor GRPO. The results confirm our theoretical analysis, demonstrating that for a Gaussian Mixture Model like DiffusionDrive, which partitions the action space by driving intentions, performing advantage estimation within each anchor is critically important.
\input{tables/abl_intra}

\paragraph{Impact of Inter-Anchor Truncated GRPO.} Tab.~\ref{tab:abl2} shows the impact of Inter-Anchor Truncated GRPO. Without this component, the model is unable to perform cross-mode comparisons, as the advantage estimates lose their global comparability and provide a misleading learning signal. In contrast, our method introduces global information for cross-mode evaluation, leading to superior performance.
\input{tables/abl_2}

%% file: tables/navsim.tex
\begin{table*}[h]
\centering
\resizebox{0.9\textwidth}{!}{
\begin{tabular}{l|c|cccccc}
    \toprule
    Method & Img. Backbone & NC $\uparrow$ &DAC $\uparrow$ & TTC $\uparrow$& Comf. $\uparrow$ & EP $\uparrow$ & \cellcolor{gray!30}PDMS $\uparrow$  \\
    \midrule
    PARA-Drive~\cite{weng2024drive} & ResNet-34 & 97.9 & 92.4 & 93.0 & 99.8 & 79.3 & \cellcolor{gray!30}84.0 \\
    VADv2~\cite{chen2024vadv2} & ResNet-34 & 97.2 & 89.1 & 91.6 & \textbf{100} & 76.0 & \cellcolor{gray!30}80.9 \\
    UniAD~\cite{hu2023planning} & ResNet-34 & 97.8 & 91.9 & 92.9 & \textbf{100} & 78.8 & \cellcolor{gray!30}83.4 \\
    Transfuser~\cite{chitta2022transfuser} & ResNet-34 & 97.7 & 92.8 & 92.8 & \textbf{100} & 79.2 & \cellcolor{gray!30}84.0 \\
    DRAMA~\cite{yuan2024drama} & ResNet-34 & 98.0 & 93.1 & 94.8 & \textbf{100} & 80.1 & \cellcolor{gray!30}85.5 \\
    Hydra-MDP*~\cite{li2024hydra} & ResNet-34 & 98.3&96.0&94.6&\textbf{100}&78.7&\cellcolor{gray!30}86.5 \\
    Hydra-MDP++*~\cite{li2025hydra} & ResNet-34 & 97.6&96.0&93.1&\textbf{100}&80.4&\cellcolor{gray!30}86.6 \\
    GoalFlow*~\cite{xing2025goalflow} & ResNet-34 & 98.3 & 93.8 & 94.3 & \textbf{100} & 79.8 & \cellcolor{gray!30}85.7 \\
    ARTEMIS~\cite{feng2025artemis} & ResNet-34 & 98.3 & 95.1 & 94.3 & \textbf{100} & 81.4 & \cellcolor{gray!30}87.0 \\
    DiffusionDrive~\cite{liao2025diffusiondrive} & ResNet-34 & 98.2 & 96.2 & 94.7 & \textbf{100} & 82.2 & \cellcolor{gray!30}88.1 \\
    WoTE~\cite{li2025end} & ResNet-34 & \textbf{98.5} & 96.8 & \textbf{94.9} & 99.9 & 81.9 & \cellcolor{gray!30}88.3 \\
    DIVER~\cite{song2025breaking} & ResNet-34 & \textbf{98.5} & 96.5 & \textbf{94.9} & \textbf{100} & 82.6 & \cellcolor{gray!30}88.3 \\
    DriveSuprim~\cite{yao2025drivesuprim} & ResNet-34 & 97.8 & 97.3 & 93.6 & \textbf{100} & 86.7 & \cellcolor{gray!30}89.9 \\    
    \midrule
    Hydra-MDP~\cite{li2024hydra} & V2-99 & 98.4 & 97.8 & 93.9 & \textbf{100} & 86.5 & \cellcolor{gray!30}90.3 \\    
    GoalFlow~\cite{xing2025goalflow} & V2-99 & 98.4 & \textbf{98.3} & 94.6 & \textbf{100} & 85.0 & \cellcolor{gray!30}90.3 \\
    \midrule
    \thename{} (Ours) & ResNet-34 & 98.3  & 97.9  & 94.8  & 99.9  & \textbf{87.5}  & \cellcolor{gray!30}\textbf{91.2}\\
    \bottomrule
\end{tabular}%
}
\caption{Comparison on NAVSIM v1 \texttt{navtest} split with closed-loop metrics. *For fair comparison, we use the official scores of versions with the same ResNet-34 backbone.}
\vspace{-0.5cm}
\label{tab:navsim}
\end{table*}

%% file: tables/navsimv2.tex
\begin{table*}[h]
\centering
\resizebox{0.9\textwidth}{!}{
\begin{tabular}{l|cccccccccc}
    \toprule
    Method & NC $\uparrow$ & DAC $\uparrow$ & DDC $\uparrow$ & TL $\uparrow$ & EP $\uparrow$ & TTC $\uparrow$ & LK $\uparrow$ & HC $\uparrow$ & EC $\uparrow$ & \cellcolor{gray!30}EPDMS $\uparrow$ \\
    \midrule
    Ego Status MLP & 93.1 & 77.9 & 92.7 & 99.6 & 86.0 & 91.5 & 89.4 & 98.3 & 85.4 & \cellcolor{gray!30}64.0 \\
    Transfuser~\cite{chitta2022transfuser} & 96.9 & 89.9 & 97.8 & 99.7 & 87.1 & 95.4 & 92.7 & 98.3 & 87.2 & \cellcolor{gray!30}76.7 \\
    Hydra-MDP++~\cite{li2025hydra} & 97.2 & 97.5 & 99.4 & 99.6 & 83.1 & 96.5 & 94.4 & 98.2 & 70.9 & \cellcolor{gray!30}81.4 \\
    DriveSuprim~\cite{yao2025drivesuprim} & 97.5 & 96.5 & 99.4 & 99.6 & 88.4 & 96.6 & 95.5 & 98.3 & 77.0 & \cellcolor{gray!30}83.1 \\
    ARTEMIS~\cite{feng2025artemis} & 98.3 & 95.1 & 98.6 & 99.8 & 81.5 & 97.4 & 96.5 & 98.3 & - & \cellcolor{gray!30}83.1 \\
    \midrule
    \thename{} (Ours) & 97.7 & 96.6 & 99.2 & 99.8 & 88.9 & 97.2 & 96.0 & 97.8 & 91.0 & \cellcolor{gray!30}\textbf{85.5} \\
    \bottomrule
\end{tabular}%
}
\caption{Comparison on NAVSIM v2 \texttt{navtest} split with extended closed-loop metrics.}
\vspace{-0.1cm}
\label{tab:navsimv2}
\end{table*}

%% file: tables/div.tex
\begin{table}[h]
\centering
\resizebox{1.0\linewidth}{!}{
\begin{tabular}{l|cccc}
    \toprule
    Method  & $Div.$ & PDMS@1 & PDMS@5 & PDMS@10 \\
    \midrule
    $\text{Transfuser}_\text{TD}$~\cite{liao2025diffusiondrive} & 0.1 & 85.7 & 85.7 & 85.7 \\
    DiffusionDrive~\cite{liao2025diffusiondrive} & 42.3 & 93.5 & 84.3 & 75.3 \\
    \midrule
    \thename{} (Ours) & 30.3 & 94.9 & 91.1 & 84.4 \\
    \bottomrule
\end{tabular}%
}
\caption{Comparison of Diversity and Top-K PDMS for the raw generated trajectories on \texttt{navtest}. \textbf{PDMS@K} denotes the PDMS score evaluated on the Top-K ranked trajectories.}
\vspace{-0.4cm}
\label{tab:div}
\end{table}

%% file: tables/abl_1.tex
\begin{table}[h]
\centering
\resizebox{\columnwidth}{!}{%
\begin{tabular}{c|cccccc}
    \toprule
    Noise Type & NC $\uparrow$ &DAC $\uparrow$ & TTC $\uparrow$& Comf. $\uparrow$ & EP $\uparrow$ & \cellcolor{gray!30}PDMS $\uparrow$  \\
    \midrule
    Add. & \textbf{98.1}  & 97.2  & 94.4  & \textbf{99.9}  & 85.3  & \cellcolor{gray!30}89.7\\
    Multi. & \textbf{98.1}  & \textbf{97.6}  & \textbf{94.5}  & \textbf{99.9}  & \textbf{85.7}  & \cellcolor{gray!30}\textbf{90.1}\\
    \bottomrule
\end{tabular}%
} %
\caption{Comparison of different type of exploration noise, where Add. and Multi. denote Additive Noise and Multiplicative Noise, respectively.}
\label{tab:abl1}
\end{table}

%% file: tables/abl_intra.tex
\begin{table}[h]
\centering
\resizebox{\columnwidth}{!}{%
\begin{tabular}{c|cccccc}
    \toprule
    Intra-Anchor & NC $\uparrow$ &DAC $\uparrow$ & TTC $\uparrow$& Comf. $\uparrow$ & EP $\uparrow$ & \cellcolor{gray!30}PDMS $\uparrow$  \\
    \midrule
    \xmark & 97.9  & 97.3  & 93.8  & 99.5  & 84.9  & \cellcolor{gray!30}89.2\\
    \cmark & \textbf{98.1}  & \textbf{97.6}  & \textbf{94.5}  & \textbf{99.9}  & \textbf{85.7}  & \cellcolor{gray!30}\textbf{90.1}\\
    \bottomrule
\end{tabular}%
}
\caption{Impact of Intra-Anchor GRPO.}
\label{tab:abl_intra}
\end{table}

%% file: tables/abl_2.tex
\begin{table}[h]
\centering
\resizebox{\columnwidth}{!}{%
\begin{tabular}{c|cccccc}
    \toprule
    Inter-Trunc. & NC $\uparrow$ &DAC $\uparrow$ & TTC $\uparrow$& Comf. $\uparrow$ & EP $\uparrow$ & \cellcolor{gray!30}PDMS $\uparrow$  \\
    \midrule
    \xmark & 97.7  & 97.3  & 93.6  & \textbf{99.9}  & \textbf{85.7}  & \cellcolor{gray!30}89.5\\
    \cmark & \textbf{98.1}  & \textbf{97.6}  & \textbf{94.5}  & \textbf{99.9}  & \textbf{85.7}  & \cellcolor{gray!30}\textbf{90.1}\\
    \bottomrule
\end{tabular}%
}
\caption{Impact of Inter-Anchor Truncated GRPO. Inter-Trunc. stands for Inter-Anchor Truncated GRPO.}
\vspace{-0.5cm}
\label{tab:abl2}
\end{table}

%% file: sec/6_conclusion.tex
\section{Conclusion}
\label{sec:conclusion}
In this work, we have presented \thename{}. By combining our proposed Intra-Anchor GRPO and Inter-Anchor Truncated GRPO with a scale-adaptive multiplicative exploration noise, our framework resolves the dilemma between diversity and consistent high quality for DiffusionDrive, which is caused by the incomplete multi-modal supervision from its imitation learning paradigm. Comprehensive experiments and qualitative comparisons validate that \thename{} achieves the best trade-off between consistent high planning quality and mode diversity and delivers state-of-the-art closed-loop performance.

%% file: sec/X_suppl.tex
\clearpage
\setcounter{page}{1}
\maketitlesupplementary

\section{Further Experiment Settings}
\paragraph{\textbf{Dataset and Metrics}} NAVSIM v1 evaluates each trajectory by feeding it into a simulator, which then provides a score based on the trajectory's interaction with the environment. The PDM score (PDMS) serves as the closed-loop planning metric and is calculated as:
\begin{equation}
\mathrm{PDMS}=N C \times D A C \times\left(\frac{5 \times E P+5 \times T T C+2 \times C}{12}\right),
\end{equation}
where the sub-metrics NC, DAC, TTC, C, EP represent the No At-Fault Collisions, Drivable Area Compliance, Time to Collision, Comfort, and Ego Progress.

NAVSIM v2 proposes an Extended PDM score (EPDMS), based on the previous formulation, which can be expressed as:
\begin{align}
&\mathrm{EPDMS} ={}  \mathrm{NC} \times \mathrm{DAC} \times \mathrm{DDC} \times \mathrm{TL} \times \nonumber \\ 
& \frac{(5 \times \mathrm{TTC}+2 \times \mathrm{C}+5 \times \mathrm{EP}+5 \times \mathrm{LK}+5 \times \mathrm{EC})}{22}  
\end{align}
The extended sub-metrics DDC, TL, LK, HC, and EC correspond to the Driving Direction Compliance, Traffic Lights Compliance, Lane Keeping, History Comfort and Extended Comfort, respectively.

\paragraph{\textbf{Training Details}} Our training process is primarily divided into two stages. In the first stage, we train the model using reinforcement learning. We use AdamW with a learning rate of $2 \times 10^{-4}$, weight decay of $1 \times 10^{-4}$, and a cosine learning rate schedule with a 10\% linear warmup. We train for 10 epochs with a batch size of 512. Additionally, we set the minimum standard deviation of the multiplicative exploration noise to $0.04$ to push the model towards sufficient exploration and prevent entropy collapse. Furthermore, we set the standard deviation to be at least $0.1$ when evaluating the Gaussian likelihood $\log \pi_{\theta}\left(\tau_{t-1}^{k,i} \mid \tau_{t}^{k,i}\right)$, which improves training stability by avoiding large magnitude gradients. The denoising discount factor $\gamma$ is set to 0.8 to downweight the contribution of earlier noisy denoising steps in the policy gradient. 

In the second stage, we train a mode selector to select the most goal-aligned trajectory from a set of multi-modal trajectories. The learning rate and batch size are the same as in the first stage. We train for 20 epochs with a batch size of 512. To enhance the robustness of our selector, we employ data augmentation during its training phase. Specifically, we augment the trajectories from our model by applying multiplicative Gaussian exploration noise, sampled from a standard deviation range of 0.1-0.2. Furthermore, we supplement the training data by randomly sampling 1\% of trajectories from the fixed GTRS~\cite{li2025generalized} trajectory vocabulary, which further improves the selector's robustness. The detailed hyperparameters are shown in Tab.~\ref{tab:hyperparams}.
\input{tables/hyperpara}

\section{Further Ablation Studies}
\paragraph{Effect of designs in mode selector.} Tab.~\ref{tab:abl3} shows the effectiveness of our design choices in the mode selector. We observe that a more refined selector design can yield modest performance gains. Following DriveSuprim, we implement a two-stage, coarse-to-fine selector. A coarse selector first identifies the top-k candidate trajectories, which are then passed to a fine-grained selector for a more detailed comparison. This two-stage process leads to a 0.2 PDMS improvement over a single-stage approach. The auxiliary ranking loss further enhances the model's capability to discern fine-grained differences in the continuous metrics, contributing an additional 0.2 PDMS gain.
\input{tables/abl_3}
\paragraph{Impact of mode selector.} \thename{} employs a more complex mode selector compared to DiffusionDrive's classifier. To verify that our model's superiority is not primarily attributed to this more complex selector, we performed an ablation study, with results presented in Tab.~\ref{tab:abl_selector}. Equipped with the same mode selector as \thename{}, DiffusionDrive improved by 1 PDMS, but still lags significantly behind \thename{}.
\input{tables/abl_selector}
\section{Qualitative Comparison}
In this section, we provide additional qualitative comparisons of Vanilla Diffusion, DiffusionDrive, and \thename{} in challenging scenarios from the \texttt{navtest} split of the planning-oriented NAVSIM dataset~\cite{dauner2024navsim}. For each method, we generate 20 trajectories. We use highly transparent colors to represent the candidate trajectories and highlight the Top-1 and Top-10 trajectories.
\paragraph{Going straight scenarios.}
Fig.~\ref{fig:straight} illustrates the performance of different methods in relatively simple straight-driving scenarios. Vanilla Diffusion collapses to a single trajectory. While DiffusionDrive performs well in the straight-driving task, it still generates numerous colliding and off-road trajectories (circled in \textcolor{red}{red}). In contrast, the trajectories produced by \thename{} are more focused, with all candidate trajectories being of high quality.

\paragraph{Turning scenarios.}
Fig.~\ref{fig:turning} illustrates the performance of different methods in turning scenarios. Similarly, Vanilla Diffusion collapses to a single trajectory. For DiffusionDrive, while the Top-1 trajectory performs well, the Top-10 trajectory exhibits delayed turning, resulting in a collision; moreover, it generates numerous colliding candidate trajectories. In contrast, \thename{} ensures high quality across all generated trajectories while preserving diversity. For instance, in Fig.~\ref{fig:left_comp_1}, the Top-1 trajectory adopts a more conservative car-following behavior, whereas the Top-10 trajectory executes a more aggressive overtaking maneuver on the curve.

\begin{figure*}[ht!]
    \centering
    \begin{subfigure}[b]{0.85\linewidth}
        \includegraphics[width=\linewidth]{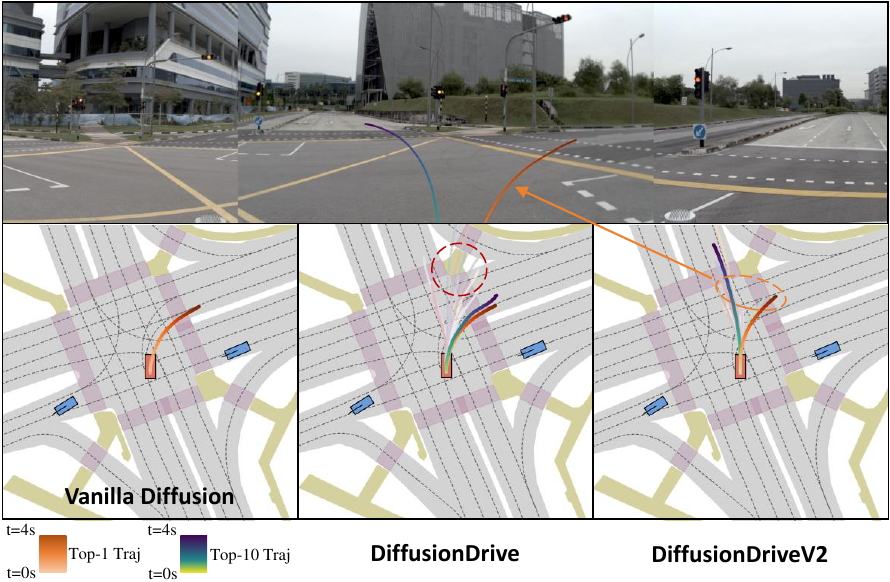}
        \vspace{-0.6cm}
        \caption{Multi-Modal Scenario 1.}
        \label{fig:multi_comp_0}
    \end{subfigure}
    \begin{subfigure}[b]{0.85\linewidth}
        \includegraphics[width=\linewidth]{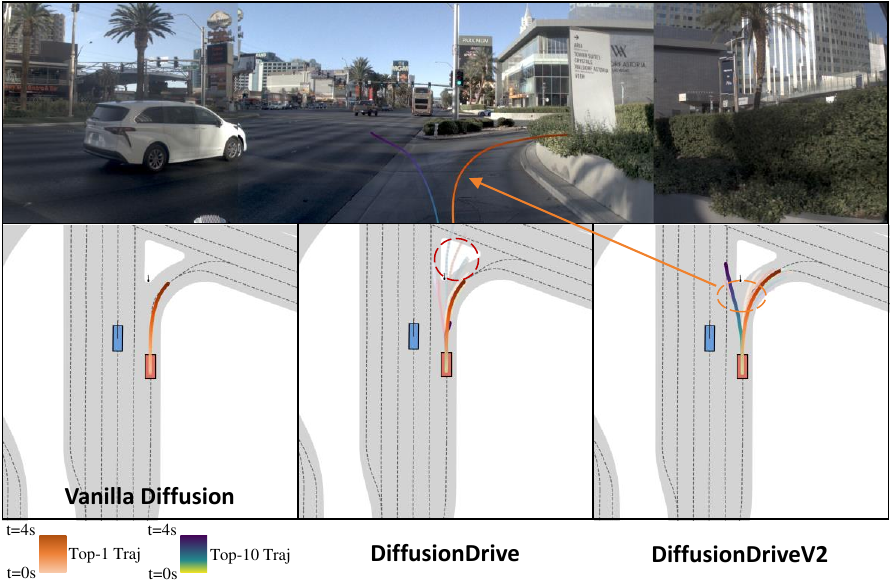}
        \vspace{-0.6cm}
        \caption{Multi-Modal Scenario 2.}
        \label{fig:multi_comp_1}
    \end{subfigure}
    \caption{\textbf{Qualitative comparison of Vanilla Diffusion, DiffusionDrive, and \thename{} on multi-modal scenarios of NAVSIM \texttt{navtest} split.}}
    \label{fig:multi_modal}
\end{figure*}

\paragraph{Multi-modal scenarios.}
To evaluate the capability of each method in handling complex driving scenarios, we conducted visualizations in scenarios with multiple potential solutions, primarily at intersections, as shown in Fig.~\ref{fig:multi_modal}. Vanilla Diffusion still suffers from mode collapse, failing to provide alternative solutions. While DiffusionDrive offers diverse trajectory options (e.g., turning vs. going straight), it fails to guarantee consistent high quality. In contrast, \thename{} effectively resolves the dilemma between diversity and consistent high quality.

\begin{figure*}[ht!]
    \centering
    \begin{subfigure}[b]{0.85\linewidth}
        \centering
        \includegraphics[width=\linewidth]{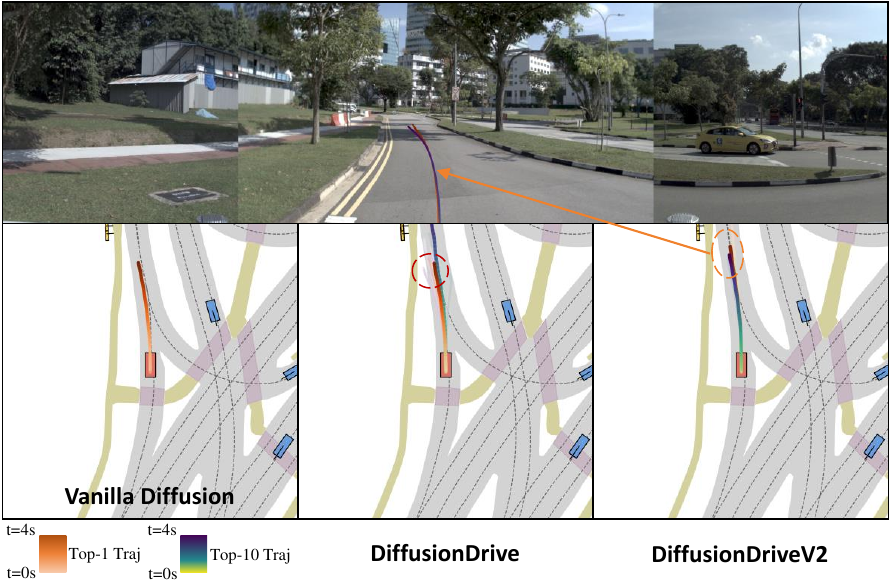}
        \vspace{-0.6cm}
        \caption{Straight Scenario 1.}
        \label{fig:straight_comp_0}
    \end{subfigure}
    
    \begin{subfigure}[b]{0.85\linewidth}
        \centering
        \includegraphics[width=\linewidth]{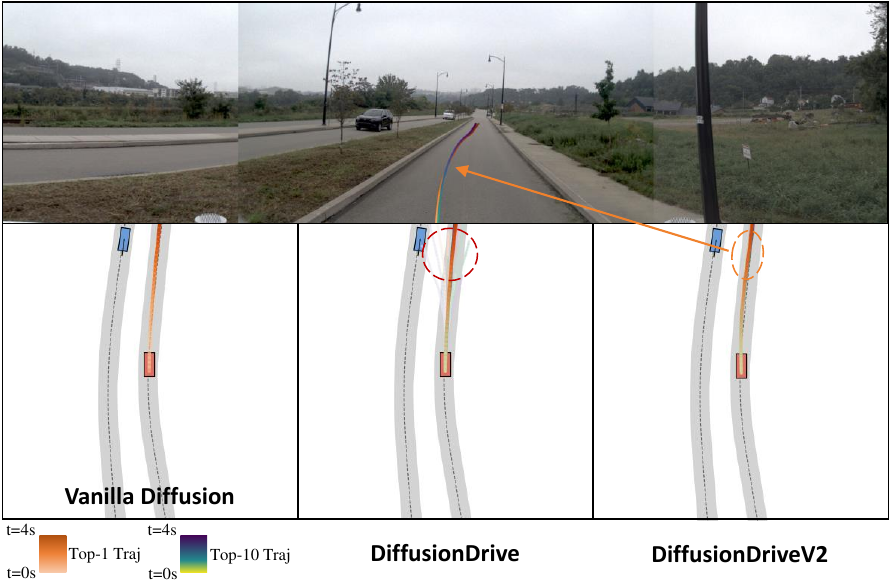}
        \vspace{-0.6cm}
        \caption{Straight Scenario 2.}
        \label{fig:straight_comp_1}
    \end{subfigure}
    
    \caption{\textbf{Qualitative comparison of Vanilla Diffusion, DiffusionDrive, and \thename{} on going straight scenarios of NAVSIM \texttt{navtest} split.}}
    \label{fig:straight}
\end{figure*}

\begin{figure*}[ht!]
    \centering
    \ContinuedFloat %
    
    \begin{subfigure}[b]{0.85\linewidth}
        \centering
        \includegraphics[width=\linewidth]{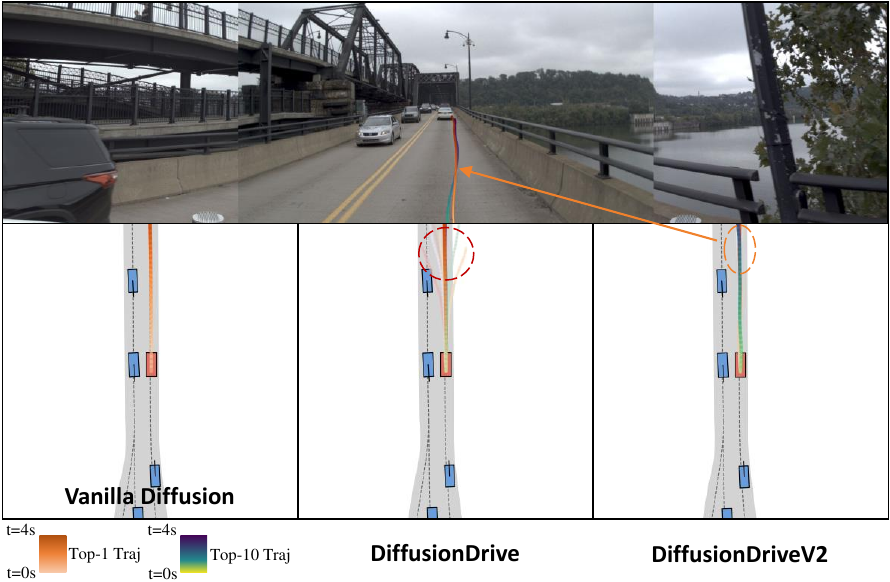}
        \vspace{-0.6cm}
        \caption{Straight Scenario 3.}
        \label{fig:straight_comp_2}
        \vspace{-0.4cm}
    \end{subfigure}
    
    \caption{\textbf{Qualitative comparison of Vanilla Diffusion, DiffusionDrive, and \thename{} on going straight scenarios of NAVSIM \texttt{navtest} split.}}
\end{figure*}
\begin{figure*}[ht!]
    \centering
    
    \begin{subfigure}[b]{0.85\linewidth}
        \centering
        \includegraphics[width=\linewidth]{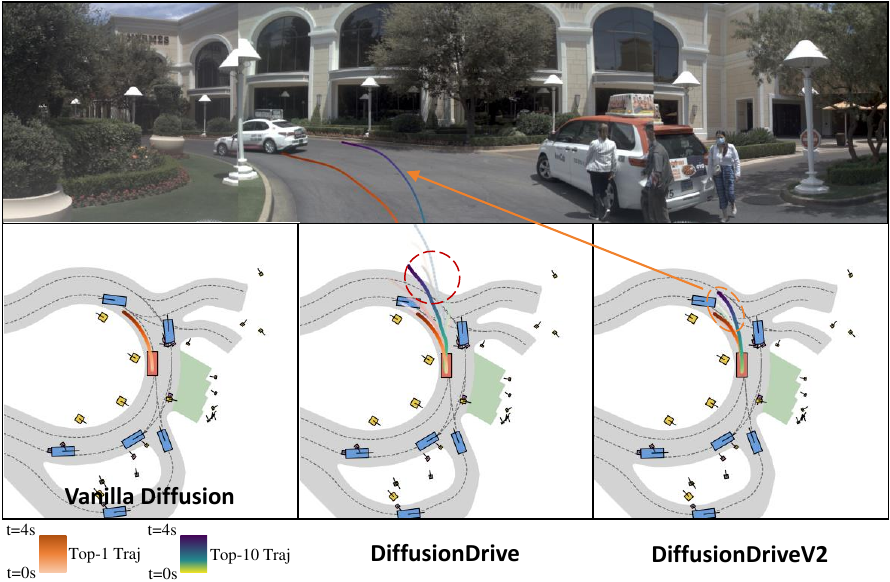}
        \vspace{-0.6cm}
        \caption{Turning Scenario 1.}
        \label{fig:left_comp_0}
    \end{subfigure}
    
    \caption{\textbf{Qualitative comparison of Vanilla Diffusion, DiffusionDrive, and \thename{} on turning scenarios of NAVSIM \texttt{navtest} split.}}
    \label{fig:turning}
\end{figure*}

\begin{figure*}[ht!]
    \centering
    \ContinuedFloat %
    
    \begin{subfigure}[b]{0.85\linewidth}
        \centering
        \includegraphics[width=\linewidth]{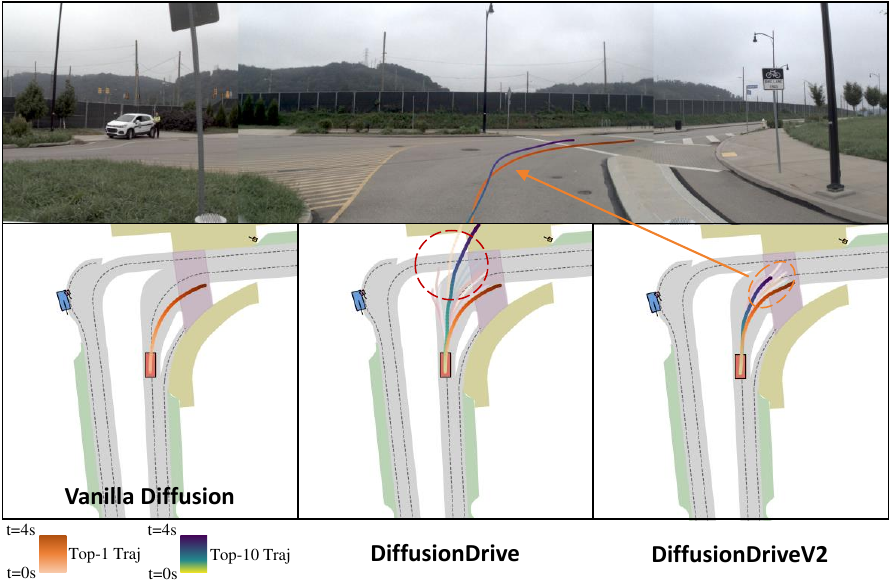}
        \vspace{-0.6cm}
        \caption{Turning Scenario 2.}
        \label{fig:left_comp_1}
    \end{subfigure}
    
    \begin{subfigure}[b]{0.85\linewidth}
        \centering
        \includegraphics[width=\linewidth]{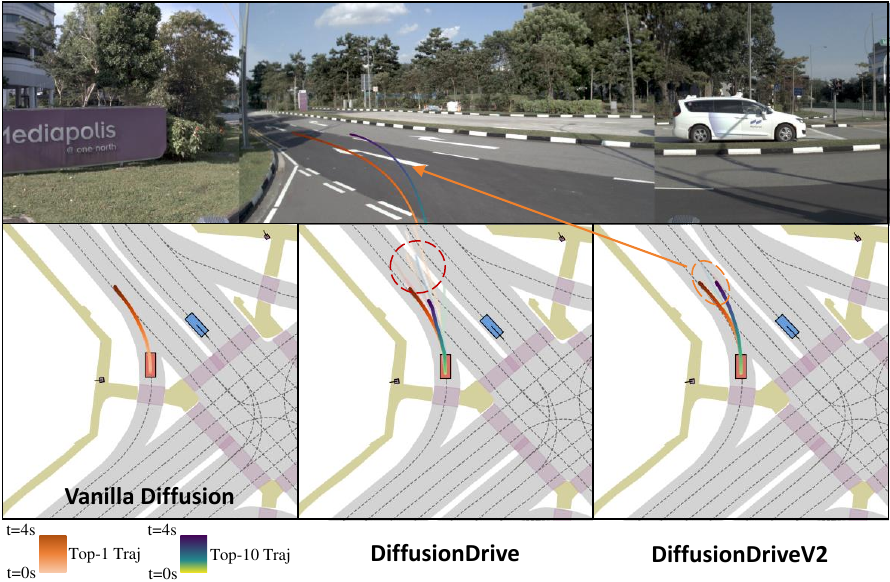}
        \vspace{-0.6cm}
        \caption{Turning Scenario 3.}
        \label{fig:left_comp_2}
        \vspace{-0.4cm}
    \end{subfigure}
    
    \caption{\textbf{Qualitative comparison of Vanilla Diffusion, DiffusionDrive, and \thename{} on turning scenarios of NAVSIM \texttt{navtest} split.}}
\end{figure*}

\newpage

%% file: tables/hyperpara.tex
\begin{table}[ht]
\centering
\caption{Hyperparameters for \thename{}.}
\label{tab:hyperparams}
\begin{tabular}{lll}
\toprule
\textbf{Stage} & \textbf{Parameter} & \textbf{Value} \\
\midrule
\multirow{10}{*}{I} & Number of epochs & 10 \\
 & Batch size & 512 \\
 & Learning rate & $2e^{-4}$ \\
 & Weight decay & $1e^{-4}$ \\
 & Warmup ratio & 0.10 \\
 & Learning rate schedule & Cosine \\
 & BC loss weight & 0.1 \\
 & Minimum denoising std & 0.04 \\
 & Minimum log-variance std & 0.10 \\
 & Discount factor & 0.8 \\
\midrule
\multirow{9}{*}{II} & Number of epochs & 20 \\
 & Batch size & 512 \\
 & Learning rate & $2e^{-4}$ \\
 & Weight decay & $1e^{-4}$ \\
 & Warmup ratio & 0.10 \\
 & Learning rate schedule & Cosine \\
 & Number of aug trajs. & 2 \\
 & Aug noise std & (0.1, 0.2) \\
 & Percentage of GTRS vocab & 1\% \\
\bottomrule
\end{tabular}
\end{table}

%% file: tables/abl_3.tex
\begin{table}[h]
\centering
\resizebox{\columnwidth}{!}{%
\begin{tabular}{llcccccc}
    \toprule
    Model & Description & NC $\uparrow$ &DAC $\uparrow$ & TTC $\uparrow$& Comf. $\uparrow$ & EP $\uparrow$ & \cellcolor{gray!30}PDMS $\uparrow$  \\
    \midrule
    $\mathcal{M}_{0}$&Base Model& 97.9  & 97.4  & 94.1  & \textbf{99.9}  & 85.4  & \cellcolor{gray!30}89.7\\
    $\mathcal{M}_{1}$&$\mathcal{M}_{0} + $Coarse2Fine& 98.0  & 97.4  & 94.3  & \textbf{99.9}  & 85.5  & \cellcolor{gray!30}89.9\\
    $\mathcal{M}_{2}$& $\mathcal{M}_{1} + $Rank Loss& \textbf{98.1}  & \textbf{97.6}  & \textbf{94.5}  & \textbf{99.9}  & \textbf{85.7}  & \cellcolor{gray!30}\textbf{90.1}\\
    \bottomrule
\end{tabular}%
}
\caption{Ablation for design choices of mode selector.}
\label{tab:abl3}
\end{table}

%% file: tables/abl_selector.tex
\begin{table}[h]
\centering
\resizebox{\columnwidth}{!}{%
\begin{tabular}{lccccccc}
    \toprule
    Model & Selector & NC $\uparrow$ &DAC $\uparrow$ & TTC $\uparrow$& Comf. $\uparrow$ & EP $\uparrow$ & \cellcolor{gray!30}PDMS $\uparrow$  \\
    \midrule
    DiffusionDrive&\xmark& 98.2  & 96.2  & 94.7  & \textbf{100}  & 82.2  & \cellcolor{gray!30}88.1\\
    DiffusionDrive&\cmark& 97.2  & 97.0  & 92.2  & \textbf{100}  & 86.8  & \cellcolor{gray!30}89.1\\
    \thename{}&\cmark& \textbf{98.3}  & \textbf{97.9}  & \textbf{94.8}  & 99.9  & \textbf{87.5}  & \cellcolor{gray!30}\textbf{91.2}\\
    \bottomrule
\end{tabular}%
}
\caption{The impact of mode selector.}
\label{tab:abl_selector}
\end{table}